\documentclass[sigconf, nonacm]{acmart}
\usepackage{booktabs}
\usepackage{multirow}
\usepackage{algorithm,algpseudocode}
\usepackage{xspace}
\usepackage{subfigure}
\usepackage{enumitem}
\usepackage{tabularx}
\usepackage{array}
\usepackage{tikz}
\usepackage{pgfplots}
\usepackage{dsfont} %% Added to support \mathds{1} indicator function
\pgfplotsset{compat=1.17}
\usetikzlibrary{arrows.meta,positioning,shapes.geometric,fit,backgrounds,calc,shadows.blur}

\newcommand{\name}{\textsc{CHAMP}\xspace}
\newcommand{\cupidv}{\textsc{Cupid}\xspace}

\newcommand{\dawn}{\textsc{Dawn}\xspace}

\title{\name: Cross-domain Hybrid Architecture for Matchmaking and Prediction in Online Multi-Player Games}

\author{Kai Wang}
\affiliation{%
  \institution{Independent Researcher}
  \city{Santa Clara}
  \country{United States}}
\email{wangjinjie722@gmail.com}

\author{Ge Fan}
\affiliation{%
  \institution{Independent Researcher}
  \city{Hangzhou}
  \country{China}}
\authornote{Corresponding Author.}

\email{ge.fan@outlook.com}

\author{Chaoyun Zhang}
\affiliation{%
  \institution{Independent Researcher}
  \city{Beijing}
  \country{China}}
\email{vyokky@163.com}

\author{Yuyang Jiang}
\affiliation{%
  \institution{The Hong Kong University of Science and Technology}
  \city{Hong Kong SAR}
  \country{China}}
\email{yjiang257@connect.ust.hk}

\author{Yuze Liu}
\affiliation{%
  \institution{Swinburne University of Technology}
  \city{Melbourne}
  \country{Australia}}
\authornotemark[1]

\email{yuzeliu@swin.edu.au}


\begin{document}

\sloppy 

\begin{abstract}

Multiplayer Online Battle Arena (MOBA) games rely on matchmaking to maintain competitive balance. Our prior work, CUPID, framed matchmaking as an assignment re-optimization problem and showed that a single-mode win-rate predictor can meaningfully rebalance teams. However, deploying such a system across diverse player populations exposes three practical bottlenecks: most queueing players lack sufficient in-mode match history (cold start), skill distributions shift drastically across rank tiers (distribution inconsistency), and extreme skill segments are severely data-starved.

We present CHAMP, a cross-domain matchmaking framework that resolves these deployment bottlenecks. To address data sparsity and cold starts, CHAMP replaces the target-mode-only player profile with a hybrid domain feature collection: a timestamp-ordered cross-mode short-term sequence whose slices are annotated with target-domain features, plus per-mode breakdowns of long-term, real-time and team statistics. We further propose the Domain-Aware Win-rate Network (DAWN): a Domain-aware Knowledge Extractor (DAKE) compiles target-mode attributes into learnable representations that feed Domain-Aware Temporal/Spatial/Permutation OmniNet Encoders (DATOE/DASOE/DAPOE), so that mode-conditioned representations and per-mode debiasing are learned jointly inside a single shared network. Online, one trained DAWN serves every supported mode, with per-mode position-satisfaction thresholds as the only mode-specific knob.

Offline, DAWN achieves 67.73\% win-rate prediction accuracy, outperforming all evaluated attention and sequence baselines. Online A/B tests across the entire League ladder of a large-scale MOBA game, from novice players up to the top-expert players served by Elite Mode, demonstrate consistent drops in imbalanced matches. For lower-tier players, CHAMP reduces the 5-minute kill crushing rate by up to 20.73\%.

\end{abstract}

\maketitle

\section{Introduction}

Multiplayer Online Battle Arena (MOBA) games have achieved remarkable popularity and economic success, attracting significant research interest in the gaming and AI communities \cite{mora2018moba, pramono2018matchmaking}. Games such as \emph{League of Legends} (LoL) and \emph{Dota 2} boast hundreds of millions of players worldwide \cite{lolcount, dota2ti}. A critical component in these games is the matchmaking system, which assembles teams of comparable skill levels to ensure fair, competitive, and enjoyable matches \cite{veron2014matchmaking, claypool2015surrender, chen2022matchmaking}.

The dominant industrial practice still relies on a single Matchmaking Rating (MMR) score, including ELO \cite{bradley1952rank}, TrueSkill \cite{herbrich2006trueskill}, TrueSkill2 \cite{minka2018trueskill}, or their learned variants \cite{zhang2022quickskill}, to group ten queueing players into two opposing teams. While computationally cheap, this scalar-skill view exposes several structural defects in a real MOBA: match fairness is approximated by the closeness of two team-MMR sums, which ignores intra-team position assignments and high-order player--player synergies that strongly influence match outcomes \cite{gong2020optmatch}; MMR is updated purely from win/loss outcomes, so it converges slowly for new or low-volume players and is noisy at the skill extremes; and once the ten players are grouped, MMR offers no mechanism to revisit team or role assignments, leaving many objectively imbalanced lobbies that the system has no further chance to repair.

To overcome these limitations, prior work \cupidv \cite{fan2024cupid} introduced a \emph{re-matchmaking} stage that intervenes after the initial MMR-based grouping. \cupidv uses a deep learning model to score candidate team-and-position permutations and pick the assignment that is most satisfying to the players and most fair as a match-up; deployed in the League mode of a popular MOBA game, it yielded significant gains in both position satisfaction and game fairness over MMR-only matchmaking. The win-rate predictor that drives this ranking, however, is trained \emph{within a single game mode}, on features collected exclusively from that mode. In a live MOBA, players rarely live inside one mode: a typical player warms up a new champion in casual or quick-match queues, tries variant lineups in 3v3 or ARAM-style brawls, and then brings the same champion into the League mode, so behavior signals from neighboring modes carry strong information about how that player will perform in the target mode. At the same time, training a dedicated re-matchmaking framework for every mode, each requiring its own data pipeline, training schedule, and serving footprint, would be operationally heavy and wasteful, especially because most modes are too data-thin to support a well-calibrated predictor on their own. Generalizing \cupidv to this realistic, multi-mode setting therefore exposes three practical challenges that motivate the present work:

\begin{enumerate}[leftmargin=*]
    \item \textbf{Player Cold Start.} A large fraction of queueing players have very limited match history in the target mode, e.g., hundreds of casual games but only a handful of League matches this season, or a recently switched main role. In-target-mode features alone yield high-variance win-rate estimates and unstable assignment decisions.

    \item \textbf{Distribution Inconsistency.} Players' behavior differs substantially across modes: 5v5 League rewards macro rotations, 3v3 brawls emphasize mechanical dueling, and ARAM compresses gold curves. A predictor trained on pooled multi-mode data without explicit mode conditioning is systematically miscalibrated on any single mode, directly biasing the fairness ranking.

    \item \textbf{Data Sparsity.} Some modes simply lack enough matches for accurate single-mode training. The most acute case is \emph{Elite Mode} (top-expert players), where the player pool is small and a full season yields orders of magnitude fewer samples than mid-tier modes. Training a dedicated predictor overfits, while a mode-agnostic predictor is dragged toward the data-rich bulk and loses accuracy where matchmaking is hardest.
\end{enumerate}

To address these challenges, we propose \name, an enhanced re-matchmaking framework that targets cross-mode matchmaking from three complementary angles: feature design, model architecture, and deployment.
At the \emph{feature} level, we replace the single-mode behavior sequence used in \cupidv with a cross-mode sequence and further enrich each player's profile with per-mode long-term statistics and real-time signals, so that the input describes a player's state in a more complete and mode-aware fashion.
At the \emph{model} level, we propose the Domain-Aware Win-rate Network (\dawn), which extends the OwO backbone with a Domain-aware Knowledge Extractor (DAKE) and Domain-Aware Temporal/Spatial/Permutation OmniNet Encoders (DATOE, DASOE, DAPOE): DAKE compiles target-mode attributes into learnable representations that are jointly consumed by the encoders, so that mode-conditioned representations and per-mode debiasing are produced inside a single shared network rather than via separate per-mode heads.
At the \emph{deployment} level, instead of training a separate re-matchmaking system per mode, we serve all supported modes from a single \dawn model and reflect mode-specific position-preference characteristics through per-mode satisfaction thresholds in the assignment filter, so that one online service consistently handles diverse game modes.
In summary, the contributions of this paper are as follows:

\begin{enumerate}[leftmargin=*]
    \item \textbf{Cross-Mode Re-Matchmaking Framework.} We extend re-matchmaking from a single-mode pipeline to a unified cross-mode framework that treats a player's behavior across every available game mode as a first-class input. The framework consolidates cross-mode behavior sequences together with per-mode long-term statistics and real-time signals into a single, mode-aware player representation, so that one re-matchmaking system can consistently serve players whose history spans multiple modes and directly mitigate player cold-start and data sparsity at the system level.

    \item \textbf{Domain-Aware Win-rate Network (\dawn).} We propose \dawn, a prediction model that extends the OwO backbone with DAKE and three Domain-Aware encoders (DATOE, DASOE, DAPOE). DAKE compiles target-mode attributes into learnable representations that are jointly consumed by the encoders, so that mode-conditioned representations and per-mode debiasing are produced inside a single shared network, yielding 67.73\% prediction accuracy that surpasses every evaluated baseline including the OwO model from \cupidv.

    \item \textbf{Unified Multi-Mode Deployment.} We deploy \name as a single online service that drives all supported game modes from one \dawn model, with mode-specific position preferences expressed through per-mode satisfaction thresholds in the assignment filter. Online A/B tests covering every League tier from novice up to the top-expert players served by Elite Mode show consistent drops in economy and kill crushing rates, with up to 20.73\% in kill crushing rate at 5~min for the lowest tier, and \name has been deployed at scale.
\end{enumerate}

\section{Related Work}

\subsection{Matchmaking Systems in Online Games}
Online MOBA matchmaking traditionally relies on a scalar MMR, estimated by ELO \cite{bradley1952rank}, TrueSkill \cite{herbrich2006trueskill}, or TrueSkill2 \cite{minka2018trueskill}. Subsequent work goes beyond scalar skill: OptMatch \cite{gong2020optmatch} models high-order player--champion interactions, GloMatch \cite{deng2021globally} optimizes global match quality with deep reinforcement learning, and QuickSkill \cite{zhang2022quickskill} addresses cold-start skill estimation.

Orthogonal to skill estimation, win prediction models forecast match outcomes either pre-match from roster and history \cite{do2021using, gu2021neuralac, wang2020match} or in-game from live state \cite{yang2022interpretable, gu2023massne, semenov2017performance, makarov2018predicting, hodge2019win, hitar2022machine}. Bridging both threads, the \cupidv framework \cite{fan2024cupid} introduces \emph{re-matchmaking} (re-optimizing team and position assignments after the initial MMR lobby is formed) and proposes the OwO encoder atop OmniNet \cite{tay2021omninet}, with temporal, spatial, and permutation channels, as the underlying pre-match win-rate predictor.

\subsection{Domain-Aware Modeling in Data Mining}
Domain-aware modeling has emerged from two complementary lines of data mining research. The first is \emph{cross-domain transfer learning} \cite{zhu2021cross, pan2010survey, li2020ddtcdr, fan2022mv, liu2024development, fan2026uniboost, liu2026structure}, in which a data-rich source domain is used to lift performance on a cold or sparse target domain. The second is \emph{multi-domain architectures} such as STAR \cite{sheng2021star}, MMOE \cite{ma2018modeling}, and CFM \cite{fan2021predicting, fan2023improving}, which share most parameters across domains while reserving lightweight domain-specific structure for residual heterogeneity. Adjacent to both, target-relevance attention models such as DIN \cite{zhou2018din} and DIEN \cite{zhou2019dien} condition representations on a target \emph{item} rather than a target \emph{domain}, and are therefore complementary rather than substitutable.

\name inherits the multi-domain conditioning paradigm from this lineage and applies it to MOBA win prediction by treating each game mode as a domain. The Domain-Aware encoders inject a learnable representation of the target mode directly into the OwO backbone via dedicated domain context tokens, so mode-conditioned representations and per-mode debiasing are learned inside a single shared network rather than as a post-hoc reweighting layer.

\section{Preliminary}

\subsection{Background and Re-matchmaking}
A comprehensive treatment of MOBA matchmaking can be found in \cupidv \cite{fan2024cupid}; here we summarize only the components on which \name builds. Modern production matchmaking systems decompose the team-formation task into two sequential stages that operate at distinct granularities: a coarse-grained \emph{pre-matching} stage that assembles candidate lobbies from the live queue, followed by a fine-grained \emph{re-matchmaking} stage that determines the final team-and-position assignment within each lobby.

\textbf{Pre-matching} operates over the entire live player queue. It periodically picks 10 players whose MMR values fall within a narrow window and groups them into a single lobby, with the dual objectives of approximate skill parity at the player level and short queue times. This stage must scale to millions of concurrent queueing players and therefore relies on cheap, scalar MMR-window heuristics rather than per-lobby modeling.

\textbf{Re-matchmaking} then operates inside each grouped 10-player lobby. The role-and-team layout \emph{within} a lobby is a much smaller combinatorial space (a fixed set of role/team assignments over 10 players) but has a disproportionate impact on match outcome: a balanced MMR pool can still produce blowouts if positions are misassigned or if one team systematically concentrates win-rate-correlated traits. Re-matchmaking exploits this small candidate space to optimize both position satisfaction and predicted match fairness using machine-learning models that would be far too expensive to run over the full queue.

The two stages are complementary: pre-matching keeps queue times tractable at production scale, while re-matchmaking spends modeling budget exactly where it matters, namely the small set of intra-lobby permutations that actually drive blowouts. The rest of this paper focuses on the re-matchmaking stage, which proceeds in two further steps as detailed below.

\textbf{Step 1: Candidate generation by position preference.} A position-preference program enumerates a set of candidate team-and-position assignments $\mathcal{A} = \{A_{t_1}^{i_1, 1}, \ldots, A_{t_N}^{i_N, N}\}$ over the 10 grouped players, and scores each candidate by its overall position satisfaction
\begin{equation}
\mathcal{P}_\mathcal{A} = \prod_{n=1}^N p_n^{i_n},
\end{equation}
where $p_n^{i_n}$ is player $n$'s satisfaction score for position $i_n$. Candidates whose satisfaction falls below a threshold $\mathcal{P}_\tau$ are pruned, leaving only assignments whose role mapping is acceptable to all 10 players.

\textbf{Step 2: Final selection by win-rate prediction.} The surviving candidates are then re-ranked by a win-rate prediction model. For each candidate $\mathcal{A}$ the model produces a predicted win rate $\hat{\mathbf{y}}_\mathcal{A}$, from which we derive a fairness score
\begin{equation}
s_\mathcal{A} = 1 - 2 |\hat{\mathbf{y}}_\mathcal{A} - 0.5|,
\end{equation}
which peaks at $1$ when the predicted win rate is exactly $0.5$ (a perfectly balanced matchup) and decays toward $0$ as the predicted outcome becomes more lopsided. The system returns the candidate with the highest $s_\mathcal{A}$, i.e., the position assignment whose predicted win rate is closest to 50\%.

The accuracy of $\hat{\mathbf{y}}_\mathcal{A}$ is therefore the binding constraint on online fairness: a miscalibrated win-rate predictor will surface lopsided lobbies even when truly fair candidates exist in $\mathcal{A}$. This motivates our focus on cross-domain win-rate prediction in the rest of the paper.

\subsection{Cross-Domain Setup}
Modern MOBA games host heterogeneous player sub-populations that differ in skill distribution, behavior, and match volume. In this work we treat each such sub-population, whether a distinct game mode or a distinct rank tier within a mode, as a separate \emph{domain} for matchmaking. Offline experiments aggregate across game modes; online evaluation stratifies by rank tier (Table~\ref{tab:online_tier}), reflecting both senses of domain in our empirical setup.

Let $\mathcal{D} = \{d_1, d_2, \ldots, d_M\}$ denotes the set of $M$ domains available during training. For a given matchmaking request in domain $d_m$, the system predict match outcomes using potentially limited in-domain data, while leveraging cross-domain knowledge from the remaining domains $\mathcal{D} \setminus \{d_m\}$.

\section{The Design of \name}

\subsection{\name in a Nutshell}
The \name architecture keeps the original \cupidv assignment filter and re-matchmaking loop, but redesigns three pieces of the pipeline so that a single re-matchmaking system can serve every game mode of a live MOBA:
\begin{enumerate}[leftmargin=*]
    \item \textbf{Hybrid Domain Feature Collection.} Each player is represented by a hybrid input that mixes behavior across modes: a timestamp-ordered cross-mode short-term sequence in which every slice is annotated with target-domain-specific features, plus long-term, real-time, and team statistics that are additionally broken down per mode.
    \item \textbf{DAWN with Domain-Aware Encoders.} The win-rate predictor is \dawn. DAKE compiles the target mode's match-related attributes into a learnable domain representation, which is fed into the upgraded DATOE, DASOE, and DAPOE encoders so that DATOE and DASOE learn mode-conditioned per-player representations and DAPOE supplies additional match context together with an in-network debiasing channel.
    \item \textbf{Unified Online Deployment.} A single trained \dawn instance drives every supported mode online; per-mode position-satisfaction thresholds in the assignment filter are the only mode-specific knobs, so one service consistently handles all supported game modes.
\end{enumerate}

 \begin{figure*}[t] \centering \includegraphics[width=\textwidth, trim=0 80 0 80, clip]{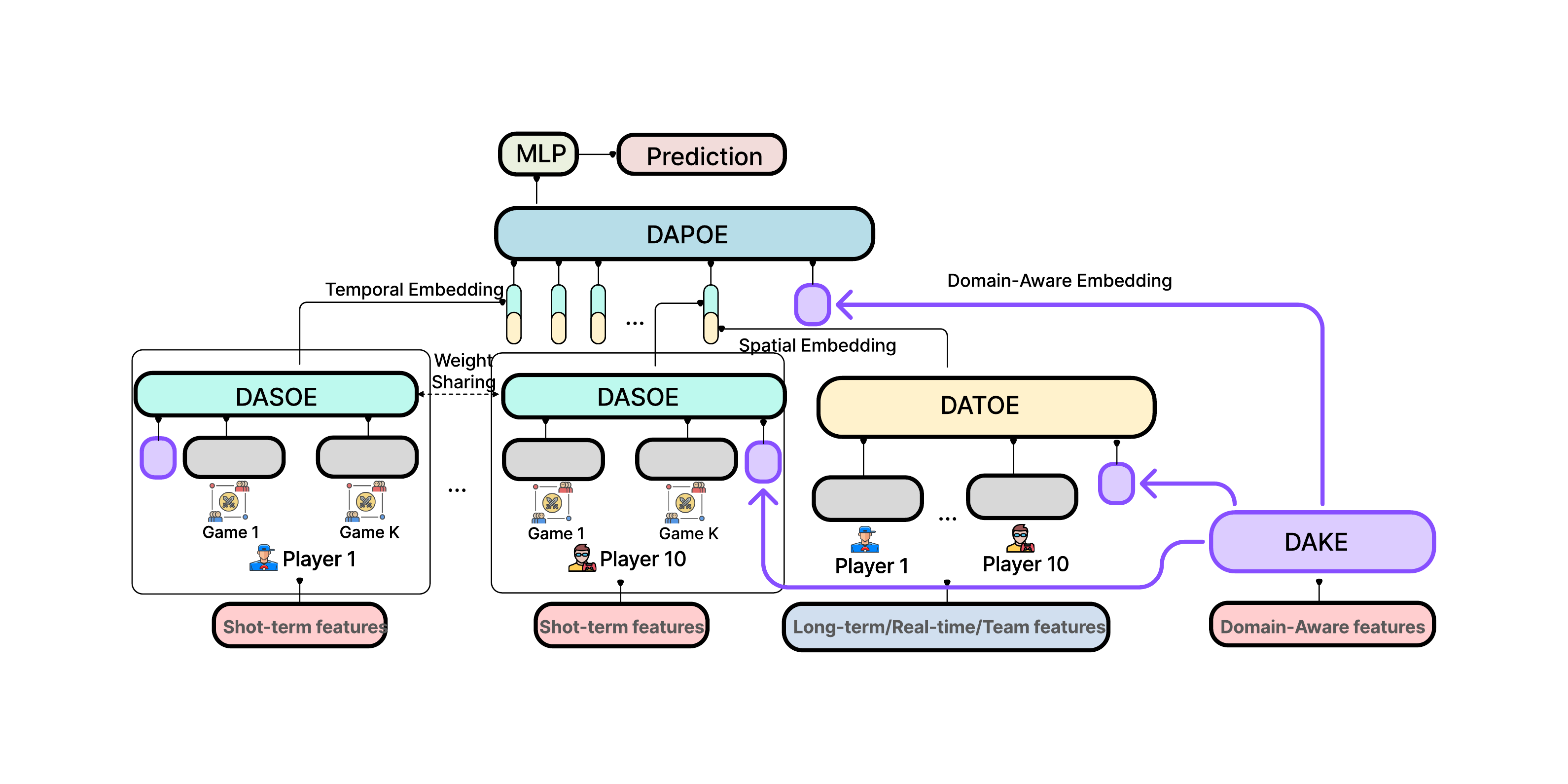} 
 \caption{Overall architecture of \name, extending \cupidv with the Hybrid Domain Feature Collection, \dawn (DAKE + DATOE/DASOE/DAPOE)} \label{fig:dawn} 
 \vspace{-4mm}

 \end{figure*}

\subsection{Hybrid Domain Feature Collection}
\label{sec:cross_feature}
A standard matchmaking pipeline characterizes each player from a single bag of in-target-mode features. \name instead constructs a \emph{hybrid} input that interleaves cross-mode behavior with target-domain-specific signals on two complementary time scales: a timestamp-ordered short-term sequence and a per-mode statistical profile. Throughout this section we treat a matchmaking request in target mode $d_m\in\mathcal{D}$ and parameterize this context with a learnable mode embedding $\mathbf{e}_m\in\mathbb{R}^{d_e}$.

\paragraph*{(i) Cross-mode short-term sequence with per-slice domain features.}
For each player $n$ we fetch the last $K$ matches across \emph{all} modes and order them by timestamp into a single sequence
\begin{equation}
\mathbf{X}^{ST}_n \;=\; \big[\,(\mathbf{f}^{(1)}_n,\, \mathbf{f}^{d}_{n,1},\, m_1),\,\ldots,\,(\mathbf{f}^{(K)}_n,\, \mathbf{f}^{d}_{n,K},\, m_K)\,\big],
\label{eq:short_term_seq}
\end{equation}
where $\mathbf{f}^{(k)}_n$ is the per-match action profile (KDA, gold, objectives, etc.), $m_k\in\mathcal{D}$ is the mode of the $k$-th match, and $\mathbf{f}^{d}_{n,k}$ is a per-slice supplement describing how that match relates to the \emph{target} domain $d_m$ (e.g.\ position/role match with the requested slot, mode-affinity statistics, recency under $d_m$). The mixed ordering preserves the natural temporal interleaving of a player's behavior, while the per-slice domain features tell the model how to compare a casual game with a League one when both are queried by the same target mode.

\paragraph*{(ii) Per-mode long-term, real-time, and team statistics.}
Beyond the recent sequence, \name also enriches the long-term, real-time, and team-level views with per-mode breakdowns. Concretely, we concatenate one statistics vector per mode together with a global summary,
\begin{equation}
\mathbf{X}^{LT}_n \;=\; \big[\,\mathbf{x}^{LT}_{n,d_1};\;\mathbf{x}^{LT}_{n,d_2};\;\ldots;\;\mathbf{x}^{LT}_{n,d_M};\;\mathbf{x}^{LT}_{n,*}\,\big],
\label{eq:long_term_feat}
\end{equation}
where $\mathbf{x}^{LT}_{n,d_j}$ aggregates the player's long-term performance, real-time form indicators and team-context statistics restricted to mode $d_j$, and $\mathbf{x}^{LT}_{n,*}$ is the corresponding mode-agnostic summary. Together, equations~(\ref{eq:short_term_seq})--(\ref{eq:long_term_feat}) provide a richer, multi-resolution view of the player than a target-mode-only profile, and serve as the raw inputs consumed by \dawn in Section~\ref{sec:dawn}.

\subsection{DAWN Architecture}
\label{sec:dawn}

The core win-rate predictor in \name is \dawn, illustrated in Figure~\ref{fig:dawn}. \dawn keeps the three-encoder OwO backbone of \cupidv~\cite{fan2024cupid} but upgrades each component into a domain-aware variant. DAKE compiles attributes of the target mode into learnable representations, which the upgraded DATOE, DASOE, and DAPOE consume jointly with their original inputs.

\subsubsection{Domain-aware Knowledge Extractor (DAKE).}
For a target mode $d_m$, DAKE takes as input the learnable mode embedding $\mathbf{e}_m$ together with a vector $\boldsymbol{\phi}_m$ of mode-related attributes (map type, party size, role/position layout, mode-level statistics, etc.) and produces a sequence of $T$ domain context tokens
\begin{equation}
\mathbf{C}_m \;=\; \mathrm{DAKE}\!\big(\mathbf{e}_m,\,\boldsymbol{\phi}_m\big) \;\in\; \mathbb{R}^{T\times d}.
\label{eq:dake_tokens}
\end{equation}
DAKE thus turns mode-side knowledge into a learnable, fixed-shape representation that downstream encoders can attend to and concatenate with their own inputs.

\subsubsection{Domain-Aware Encoders (DATOE, DASOE, DAPOE).}
Each encoder is an OmniNet~\cite{tay2021omninet}, a Transformer augmented with omnidirectional residual attention,
\begin{equation}
\mathrm{OmniNet}(\mathbf{X}) = \mathrm{Transformer}(\mathbf{X})_L + \mathrm{MLP}\!\big(O_{\text{att}}(\mathbf{X})\big),
\label{eq:omninet}
\end{equation}
where $O_{\text{att}}$ aggregates hidden states across all layers. The DA-variants accept the original encoder input and additionally take the DAKE tokens $\mathbf{C}_m$ as an extra input slot, so that mode-conditioned representations can be learned \emph{inside} the encoder rather than only as a late-stage adjustment. Concretely, for the $n$-th player in a candidate assignment $A$,
\begin{align}
\tilde{\mathbf{h}}^{T}_n &= \mathrm{DATOE}\!\big([\mathbf{X}^{ST}_n;\;\mathbf{C}_m]\big), \qquad n=1,\dots,10, \label{eq:datoe}\\
\tilde{\mathbf{h}}^{S}_n &= \mathrm{DASOE}\!\big([\mathbf{X}^{LT}_n;\;\mathbf{C}_m]\big), \qquad n=1,\dots,10, \label{eq:dasoe}
\end{align}
where $\mathbf{X}^{ST}_n$ and $\mathbf{X}^{LT}_n$ are the hybrid features defined in equations~(\ref{eq:short_term_seq})--(\ref{eq:long_term_feat}). Conditioning on $\mathbf{C}_m$ lets DATOE and DASOE adapt \emph{what} to look at on a per-mode basis: a strong KDA, a meaningful gold-per-minute, or a useful objective rotation contribute to win prediction differently across Casual , League, Elite modes.

DAPOE then aggregates the per-player encodings together with a pooled domain summary $\mathbf{c}_m=\mathrm{Pool}(\mathbf{C}_m)$:
\begin{equation}
\mathbf{z} \;=\; \mathrm{DAPOE}\!\big([\,\tilde{\mathbf{h}}^{T}_1;\tilde{\mathbf{h}}^{S}_1;\dots;\tilde{\mathbf{h}}^{T}_{10};\tilde{\mathbf{h}}^{S}_{10};\;\mathbf{c}_m\,]\big),
\label{eq:dapoe}
\end{equation}
where the player order is fixed by the team-and-position layout of $A$, so that any reassignment in re-matchmaking translates directly into a change of input order~\cite{fan2024cupid}. The dedicated $\mathbf{c}_m$ slot supplies DAPOE with explicit match-level context and also acts as an \emph{in-network debiasing channel}: gradients prefer to flow through this dedicated mode pathway rather than baking mode-specific shifts into the shared per-player representations, leaving the player encodings cleaner and more transferable across modes.

\subsubsection{Prediction Head and Training Objective.}
A single shared head $g(\cdot)$ maps the pooled representation $\mathbf{z}$ to the win probability,
\begin{equation}
\hat{y} = \sigma\!\big(g(\mathbf{z})\big),
\label{eq:head}
\end{equation}
and \dawn is trained with the Normalized Economy Difference (NED) loss inherited from \cupidv,
\begin{equation}
\mathcal{L}_{NED} = -\big(\alpha\, y\,\log\hat{y} + (1-\alpha)\,(1-y)\,\log(1-\hat{y})\big),
\label{eq:ned}
\end{equation}
where $\alpha = (TE_1 - TE_2)/\max(TE_1, TE_2)$ rescales the penalty by the normalized team-economy gap so that close, hard-to-predict games dominate the gradient signal. Because mode information is already injected via DAKE and DAPOE, no auxiliary per-mode head or post-hoc calibration term is required.

\subsection{Online Deployment}
\label{sec:deploy}
A single trained \dawn instance drives every supported game mode online. For a matchmaking request in target mode $d_m$ over a 10-player lobby, the live system proceeds in the following four steps:
\begin{enumerate}[leftmargin=*]
    \item \textbf{Feature retrieval and processing.} For each of the 10 players, the online feature store is queried to retrieve and process the hybrid domain features: the cross-mode short-term sequence $\mathbf{X}^{ST}_n$ (annotated with per-slice target-domain features) and the per-mode statistics $\mathbf{X}^{LT}_n$ (long-term, real-time, and team breakdowns).
    \item \textbf{Candidate generation by position satisfaction.} A position-preference program enumerates candidate team-and-position assignments. Each candidate $A$ is scored by its overall position-satisfaction product $\mathcal{P}_A = \prod_{n=1}^{N} p_n^{i_n}$, and only assignments whose satisfaction exceeds the per-mode threshold $\tau_m$ are retained as the candidate set $\mathcal{A}$.
    \item \textbf{Win-rate scoring.} The surviving candidates are forwarded to \dawn. For each $A \in \mathcal{A}$, DAKE builds the domain context $\mathbf{C}_m$ from the target-mode attributes, and the DATOE/DASOE/DAPOE encoders produce the match-level representation $\mathbf{z}$, from which the shared head yields the predicted win probability $\hat{y}(A)$ (Eqs.~\ref{eq:dake_tokens}--\ref{eq:head}).
    \item \textbf{Fairness-optimal assignment selection.} The candidate whose predicted win probability is closest to 50\% is returned as the final assignment, i.e., $A^* = \arg\min_{A \in \mathcal{A}}\,|\hat{y}(A) - 0.5|$.
\end{enumerate}
Because \dawn is a single shared model, deploying a new mode reduces to registering its attribute vector $\boldsymbol{\phi}_m$ and per-mode threshold $\tau_m$; no per-mode model is trained or hosted.

\section{Experiments}
\label{sec:offline}
In this section, we design experiments to answer the following three research questions:
\begin{itemize}[leftmargin=*]
    \item \textbf{RQ1.} How does \dawn compare with state-of-the-art methods on pre-match win prediction?
    \item \textbf{RQ2.} How do the proposed components (Hybrid Domain Feature Collection, DAKE, and the three Domain-Aware encoders) actually contribute to the overall model performance?
    \item \textbf{RQ3.} How does deploying \dawn in a live production matchmaking pipeline affect downstream business metrics?
\end{itemize}
RQ1 and RQ2 are answered offline in this section; RQ3 is answered through large-scale online A/B testing in Section~\ref{sec:online}.

\subsection{Experiment Settings}

\subsubsection{Datasets.}
We collect industrial match logs from a popular online MOBA title and construct three datasets corresponding to the three game modes that the production matchmaking system serves: \textbf{Casual}, \textbf{League}, and \textbf{Elite} (the dedicated mode for top-expert players). The three datasets contain approximately \textbf{40M}, \textbf{30M}, and \textbf{0.1M} samples, respectively, exhibiting a clear long-tail in data scale: Elite, which serves the smallest top-expert player pool, is also the sparsest mode and is therefore the primary stress-test for cold-start and data-sparsity behavior. To prevent any temporal leakage, we partition each dataset \emph{chronologically} rather than by random sampling: the earliest matches form the training set, the most recent form the test set, and 5\% of the training set (still earlier than the test horizon) is held out as a validation split for hyperparameter tuning and early stopping.

\subsubsection{Baselines.}
Following the protocol of \cupidv \cite{fan2024cupid}, we benchmark \dawn against: Logistic Regression (LR) \cite{schmidt2017minimizing} and a three-layer MLP \cite{khotanzad1990classification} as shallow baselines; LSTM \cite{hochreiter1997long} as a sequence baseline; and Transformer \cite{vaswani2017attention} and OwO \cite{tay2021omninet} (the omnidirectional attention backbone adopted by \cupidv) as attention baselines. We additionally include \textbf{\dawn-single}, an instance of \dawn trained on the data of a single target mode (i.e., the same single-domain regime as every baseline), to isolate the effect of multi-mode joint training from the architecture itself.

\subsubsection{Evaluation Metrics.}
Our pre-match win prediction task is naturally class-balanced (each match has one winning team and one losing team), so we report the two most widely used metrics for balanced binary prediction: \textbf{Accuracy (ACC)}, which captures the discrete classification quality, and \textbf{Root Mean Square Error (RMSE)}, which captures the calibration of the predicted win probabilities. Higher ACC and lower RMSE both indicate better performance~\cite{chen2019preference, fan2022pppne,chen2023hadamard}.

\subsubsection{Implementation Details.}
\dawn is optimized via Adam \cite{kingma2014adam} in TensorFlow \cite{abadi2016tensorflow}. To ensure a rigorously controlled comparison, the core OwO hyperparameter footprint (layer depth, hidden dimensions, attention heads) within \dawn mirrors the original OwO configuration.

\begin{table}[t]
\centering
\caption{Model comparison across three datasets (bold = best).}
\label{tab:model_comparison}
\small
\setlength{\tabcolsep}{3pt}
\begin{tabular}{l cc cc cc}
\toprule
\multirow{2}{*}{Model}
  & \multicolumn{2}{c}{Casual}
  & \multicolumn{2}{c}{League}
  & \multicolumn{2}{c}{Elite} \\
\cmidrule(lr){2-3} \cmidrule(lr){4-5} \cmidrule(lr){6-7}
  & ACC    & RMSE
  & ACC    & RMSE
  & ACC    & RMSE \\
\midrule
LR
  & 0.5657 & 0.4976
  & 0.5040 & 0.5069
  & 0.5912 & 0.4938 \\
MLP
  & 0.5753 & 0.4771
  & 0.5153 & 0.4866
  & 0.6041 & 0.4725 \\
LSTM
  & 0.5931 & 0.4582
  & 0.5231 & 0.4701
  & 0.6336 & 0.4514 \\
Transformer
  & 0.6012 & 0.4412
  & 0.5429 & 0.4518
  & 0.6528 & 0.4316 \\
OwO
  & 0.6321 & 0.4203
  & 0.5565 & 0.4349
  & 0.6559 & 0.4160 \\
DAWN-single
  & 0.6534 & 0.4109
  & 0.5622 & 0.4291
  & 0.6580 & 0.4098 \\
DAWN
  & \textbf{0.6612} & \textbf{0.4022}
  & \textbf{0.5685} & \textbf{0.4216}
  & \textbf{0.6773} & \textbf{0.3987} \\
\bottomrule
 \vspace{-6mm}

\end{tabular}
\end{table}

\subsection{Results Compared with Baselines (RQ1)}

Table~\ref{tab:model_comparison} reports pre-match win prediction performance. We include \dawn-single (trained on one target mode, same regime as all baselines) and \dawn (jointly consuming all three modes via the hybrid domain feature collection).

On every dataset, both variants dominate all baselines on ACC and RMSE. The performance ranking is consistent across all three modes, with \dawn at the top followed by \dawn-single, confirming that the gains stem from cross-mode joint training and the DAKE-conditioned encoders rather than dataset-specific tuning.

Crucially, \dawn outperforms \dawn-single on every cell, and the gap widens under data scarcity. On Elite, the sparsest dataset, \dawn-single barely edges out OwO (0.6580 vs.\ 0.6559 ACC, +0.32\%), whereas \dawn jumps to 0.6773 (+2.93\% over \dawn-single, the largest mode-wise gain). This confirms that cross-mode joint training effectively transfers behavioral knowledge from data-rich to data-sparse modes, directly mitigating the cold-start bottleneck that motivates \name.

\subsection{Ablation Studies (RQ2)}

\begin{figure}[t]
\centering
\includegraphics[width=1\linewidth]{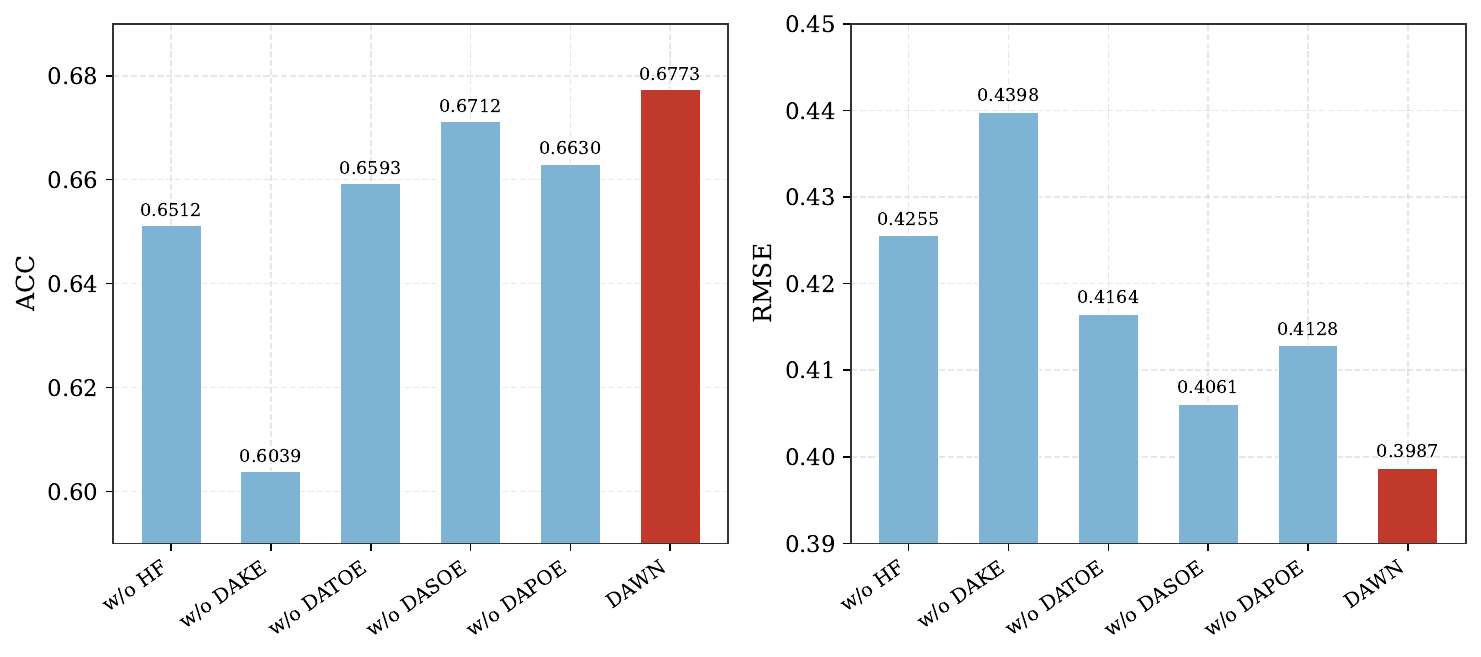}
\caption{Ablation results on the Elite dataset.}
\label{fig:ablation}
 \vspace{-3mm}
\end{figure}

Figure~\ref{fig:ablation} reports the ablation study on the sparse Elite dataset; the Casual and League datasets exhibit the same monotone ordering across all variants, and are omitted for brevity. We construct each variant by removing one component of \dawn while keeping the rest intact: \emph{w/o HF} replaces the hybrid domain feature collection with the single-mode sequence and statistics used by every baseline; \emph{w/o DAKE} drops DAKE (so the encoders are the original OwO encoders), which is equivalent to training a vanilla OwO on the union of all modes; and \emph{w/o DATOE / DASOE / DAPOE} substitutes the corresponding Domain-Aware encoder with its plain OwO counterpart.

Across all five variants the deletion of any single component produces a measurable degradation relative to the full \dawn (ACC drops of 0.61--7.34 points on Elite), and the same monotone ordering holds for RMSE. This confirms that every component of the architecture, including the hybrid feature collection, the DAKE conditioning module, and each of the three Domain-Aware encoders, contributes complementary, non-redundant capacity, and that the design as a whole is internally well-balanced rather than dominated by a single block.

Most strikingly, removing DAKE causes by far the largest collapse: ACC plummets to 0.6039, which is even \emph{below} the 0.6559 achieved by the single-mode OwO baseline on the same Elite dataset (Table~\ref{tab:model_comparison}). In other words, naively pooling all modes into a vanilla OwO does not just forfeit \dawn's gains; it actively underperforms training that vanilla OwO on Elite data alone. This indicates that without an explicit mode-conditioning signal, the heterogeneous distributions and behavioral conventions of different modes appear as noise to a shared encoder, and cross-mode supervision becomes harmful rather than beneficial. DAKE's learnable target-mode tokens are therefore the load-bearing component that makes multi-mode training viable: only once the encoders are conditioned on the target mode does aggregating cross-mode data turn from a liability into the source of \dawn's headline gains.

% \section{Online Experiments}

\subsection{Online Experiments (RQ3)}
\label{sec:online}
While offline accuracy validates the representational power of \dawn, the ultimate test of a matchmaking engine is whether it suppresses blowout matches in a live, non-stationary environment. We therefore deployed \name in two production game modes of a popular online MOBA title, specifically \textbf{League Mode} and \textbf{Elite Mode} (the dedicated mode serving top-expert players), and ran strict A/B testing against the production \cupidv system as the control. The experiment spanned over 30 days, covered tens of millions of players, and generated billions of player-game participations. All improvements reported in Tables~\ref{tab:online_overall}--\ref{tab:online_tier} are statistically significant (two-proportion $z$-test, $p < 0.01$).

Following \cupidv \cite{fan2024cupid}, we track two primary indicators of match imbalance: \textbf{Economy Crushing Rate}, the proportion of matches whose team gold differential breaches a critical threshold at 5, 10, or 15 minutes; and \textbf{Kill Crushing Rate}, the proportion exhibiting a critical kill differential at 5 and 15 minutes. Negative shifts indicate fewer crushing matches, i.e., a tighter and more competitive experience.

\begin{table}[t]
\caption{Online A/B test results. Negative is better.}
\centering
\small
\begin{tabular}{l ccc cc}
\toprule
\multirow{2}{*}{Mode} & \multicolumn{3}{c}{Economy Crushing Rate} & \multicolumn{2}{c}{Kill Crushing Rate} \\
\cmidrule(lr){2-4} \cmidrule(lr){5-6}
 & @5min & @10min & @15min & @5min & @15min \\
\midrule
League Mode             & $-$2.14\% & $-$3.05\% & $-$3.45\% & $-$9.00\% & $-$8.05\% \\
Elite Mode               & $-$3.32\% & $-$3.75\% & $-$3.84\% & $-$9.62\% & $-$8.19\% \\
\bottomrule
\end{tabular}
\label{tab:online_overall}
 \vspace{-3mm}
\end{table}

Table~\ref{tab:online_overall} reports the aggregate impact for the two deployed modes. \name uniformly reduces both crushing-rate families on \emph{every} time horizon and in \emph{both} modes: 5-minute kill blowouts drop by 9.00\% in League and 9.62\% in Elite, while 15-minute economy blowouts fall by 3.45\% and 3.84\%, respectively. The signs are consistent across all ten (mode $\times$ metric) cells, indicating that the gain is not a horizon-specific or metric-specific artifact but a structural improvement in lobby balance.

Comparing the two modes, Elite consistently outperforms League on every metric, with the largest relative gap on early-game economy crushing ($-$3.32\% vs.\ $-$2.14\%, a 55\% larger reduction at 5 minutes). We attribute this to two reinforcing mechanisms. \emph{First}, Elite is by far the sparsest of the modes we serve (\textasciitilde0.1M samples vs.\ tens of millions in League, see Section~\ref{sec:offline}); single-mode supervision on Elite alone barely converges, and our offline results (Table~\ref{tab:model_comparison}) already showed that the marginal lift from cross-mode joint training is largest on Elite. Online, this larger predictive headroom is directly cashed in through tighter assignment filtering. \emph{Second}, the top-expert players served by Elite  sit in an extremely narrow skill band where micro-advantages compound rapidly into decisive outcomes: the same incremental gain in win-probability calibration (lower RMSE) translates into a disproportionately larger reduction in critical-imbalance events, because the crushing-rate threshold is reached by a smaller residual mismatch. DAPOE's in-network debiasing channel is most useful precisely in this regime, where the residual win-probability mass is already concentrated near $0.5$ and small calibration errors are the binding constraint on online fairness.

\section{Lessons Learned}
\label{sec:lessons}
Building and deploying \name in a live multi-mode MOBA surfaced several non-obvious insights that we believe generalize to other applied cross-domain serving systems.

\textbf{Cross-Mode Data Is Not Free Without Conditioning.}
The most counter-intuitive finding from our ablation study (Figure~\ref{fig:ablation}) is that naively pooling all modes into a vanilla OwO (the \emph{w/o DAKE} variant) collapses Elite accuracy to 0.6039, which is markedly \emph{worse} than the 0.6559 achieved by the same OwO trained on Elite data alone (Table~\ref{tab:model_comparison}). In other words, simply adding cross-mode supervision is not a free improvement; without an explicit conditioning signal, the heterogeneous distributions and behavioral conventions of different modes appear as noise to a shared encoder and actively degrade target-mode performance. The practical implication is that, in a multi-domain serving system, the conditioning channel should be designed \emph{before} the corpus is scaled up: otherwise every additional domain bolted onto a shared backbone can become a liability rather than an asset.

\textbf{Cold-Start Cohorts Drive the Production Lift.}
While Table~\ref{tab:online_overall} reports the aggregate gain in each deployed mode, the most important applied finding only becomes visible once we stratify the League population by skill tier. Table~\ref{tab:online_tier} expands the League row of Table~\ref{tab:online_overall} into per-tier reductions, exposing a pronounced asymmetry in who actually benefits from \name in production.

\begin{table}[t]
\caption{League Mode A/B results by skill tier.}
\centering
\small
\begin{tabular}{l ccc cc}
\toprule
\multirow{2}{*}{Rank Tier} & \multicolumn{3}{c}{Economy Crushing Rate} & \multicolumn{2}{c}{Kill Crushing Rate} \\
\cmidrule(lr){2-4} \cmidrule(lr){5-6}
 & @5min & @10min & @15min & @5min & @15min \\
\midrule
Novice            & $-$4.98\% & $-$7.32\% & $-$8.15\% & $-$20.73\% & $-$16.45\% \\
Junior                  & $-$2.74\% & $-$3.70\% & $-$4.35\% & $-$12.74\% & $-$10.61\% \\
Senior                  & $-$1.99\% & $-$3.00\% & $-$3.59\% & $-$8.73\%  & $-$8.76\%  \\
Expert            & $-$1.71\% & $-$2.32\% & $-$2.45\% & $-$6.30\%  & $-$5.62\%  \\
\bottomrule
\end{tabular}
\label{tab:online_tier}
 \vspace{-4mm}

\end{table}

The lift is sharply skill-dependent: the novice segment absorbs a 20.73\% reduction in 5-minute kill crushing, roughly \emph{twice} the all-League average ($-$9.00\%) and more than 3$\times$ the Expert figure ($-$6.30\%). Reading down the table, the magnitude of the reduction shrinks monotonically as the tier rises (Novice $\to$ Junior $\to$ Senior $\to$ Expert), confirming that the production payoff is concentrated precisely on the players with the thinnest in-tier behavioral histories. This inverts the common deployment intuition that the most populous, data-rich tiers should yield the largest and most stable lift, and indicates that cross-domain transfer is doing real work where single-mode models cannot fit well: the gain at Novice is essentially \name imputing the missing target-mode signal from cross-mode behavior. The applied lesson is that cross-domain matchmaking methods must be evaluated cohort by cohort; aggregate lift can systematically understate the gain on the cold-start slice where the production payoff actually lives. Note further that \emph{Elite Mode}, the mode dedicated to top-expert players, sits at the opposite end of the data spectrum (the sparsest of all modes, Section~\ref{sec:offline}) and is lifted by a different mechanism (DAPOE's in-network calibration), so the two extremes of the user distribution are rescued by different parts of the same shared architecture.

\textbf{Collapse the Mode-Specific Operational Surface.}
\name deliberately collapses every mode-specific knob into one interpretable scalar, the position-satisfaction threshold $\tau_m$ in the assignment filter, and pushes all other mode dependence into shared learnable parameters (DAKE-conditioned encoders and DAPOE's debiasing slot $\mathbf{c}_m$). This was the single largest contributor to maintainability in our deployment, replacing $N$ per-mode deployment artifacts with one.

The same collapse also reshapes online serving cost. Aggregate re-matchmaking QPS is approximately conservative across modes: when a new mode launches, players \emph{migrate} into it rather than appear net new. After Elite Mode opened, Elite's per-mode QPS spiked while League decreased by a comparable amount, leaving global QPS essentially unchanged. A per-mode service would still have to be capacity-planned for its own independent peak, so deployed capacity would scale with the number of modes even though the shared workload does not. Routing all modes through one \dawn service makes capacity follow global QPS, avoiding this multiplicative overprovisioning cost. Finally, because \dawn adds only DAKE and the domain context slots atop OwO, load tests confirmed no significant difference in mean, P50, or P99 serving latency between \name and \cupidv at the same QPS.

\section{Conclusion}

This paper presented \name, a unified cross-domain re-matchmaking framework powered by \dawn to serve multiple MOBA game modes via a single shared model. \name effectively resolves cold-start and data-sparsity bottlenecks by leveraging a Domain-aware Knowledge Extractor (DAKE) for joint representation learning. Empirically, \name achieves 67.73\% offline accuracy and reduces 5-minute kill crushing by up to 20.73\% for novice players online. Key operational guidelines include: (i) cross-mode data requires domain-aware conditioning; (ii) production gains concentrate on cold-start cohorts; (iii) collapsing mode-specific configurations to a single scalar ensures scalable maintainability.

\section*{GenAI Usage Disclosure}
The authors used large language models (e.g., ChatGPT) for grammar checking and
polishing of the manuscript. All technical content, experiments, and results
are solely the work of the authors.
\bibliographystyle{ACM-Reference-Format}
\bibliography{mmk}

\end{document}